\documentclass[runningheads]{llncs}

\usepackage{eccvabbrv}

\usepackage{graphicx}
\usepackage{booktabs}
\usepackage{algorithm}
\usepackage{algpseudocode}
\usepackage{bm}
\usepackage{bbm}

\usepackage[accsupp]{axessibility}  

\usepackage{eccv}
\usepackage{hyperref}

\usepackage{orcidlink}

\begin{document}

\title{
Calibration of Network Confidence for Unsupervised Domain Adaptation Using Estimated Accuracy}

\titlerunning{UDA confidence calibration}


\author{ Coby Penso \and Jacob Goldberger\orcidlink{0000-0002-2225-1914} }

\authorrunning{C. Penso et al.}

\institute{Engineering Faculty, Bar-Ilan University\\
\email{coby.penso24@gmail.com, jacob.goldberger@biu.ac.il}}

\maketitle

\begin{abstract}
  This study addresses the problem of calibrating network confidence while adapting a model that was originally trained on a source domain to a target domain using unlabeled samples from the target domain. The absence of labels from the target domain makes it impossible to directly calibrate the adapted network on the target domain.     To tackle this challenge, we introduce a calibration procedure that relies on estimating the network's accuracy on the target domain. 
          The network accuracy is first computed on the labeled source data  and then  is modified to represent the actual accuracy of the model on the target domain.  The proposed algorithm calibrates the prediction confidence directly in the target domain by minimizing the disparity between  the estimated accuracy and the computed confidence. The experimental results show that our method significantly outperforms existing methods, which rely on importance weighting, across several standard datasets.

\end{abstract}

\section{Introduction}

Deep Neural Networks (DNN) have shown remarkable accuracy in tasks such as classification and detection when sufficient data and supervision are present. In practical applications, it is crucial for models not only to be accurate, but  also to indicate how much confidence users can have in their predictions. DNNs generate confidence scores that can serve as a rough estimate of the likelihood of correct classification, but these scores do not guarantee a match with the actual probabilities \cite{Guo2017}. Neural networks tend to be overconfident in their predictions, despite having higher generalization accuracy, due to the possibility of overfitting on negative log-likelihood loss without affecting classification error \cite{Guo2017,Balaji2017,Hein2019}. A classifier is said to be calibrated with respect to a dataset sampled from a given distribution if its predicted probability of being correct matches its true probability. 
Various methods have been introduced to address the issue of over-confidence. Network calibration can be performed in conjunction with training (see e.g. \cite{mukhoti2020calibrating, muller2019does,mixup}). Post-hoc scaling methods for calibration, such as Platt scaling \cite{Platt1999}, isotonic regression \cite{Zadrozny2002}, and temperature scaling \cite{Guo2017}, are commonly employed. These techniques apply calibration as post-processing, using a hold-out validation set to learn a calibration map that adjusts the model's confidence in its predictions to become better calibrated.

The implementation of deep learning systems on real-world problems  is hindered by the decrease in performance when a network trained on data from one domain is applied to data from a different domain where the distribution of features changes across domains (see e.g. \cite{pmlr-v139-miller21b}). This is known as the domain shift problem. In an Unsupervised Domain Adaptation (UDA) setup we assume the availability of data from the target domain but without annotation. There is a plethora of UDA methods  based on strategies
such as adversarial training methods that aim to align the distributions of the source and target domains \cite{ganin2016domain},  or self-training algorithms  based on computing pseudo labels for the target domain data \cite{zou2019confidence}.

In this study we tackle the problem of calibrating predicted probabilities when transferring a trained model from a source domain to a target domain without any given labels.
Studies show that present-day  UDA methods are prone to learning improved accuracy at the expense of deteriorated prediction confidence 
\cite{wang2020}. Calibrating the confidence of the adapted model on data from the target domain is challenging due to the coexistence of the domain gap and the lack of target labels. 
Current UDA calibration methods use the labeled validation set  from the source domain to approximate the target domain statistics in certain aspects.
Some studies \cite {salvador2021improved,tomani2021post} propose to modify the calibration set to represent a generic distribution shift.
Other methods  \cite{park2020,wang2020,pampari2020}  apply Importance Weighting (IW)
 by assigning higher weights to source examples that resemble those in the target domain.
 In practice, current methods doesn't work well and in many cases, they yield calibration results which are worse than the uncalibrated network.
The main weakness of current IW-based methods is that they use the unlabeled target data solely to train a binary source/target classifier, but the actual calibration is done on the source domain data while the target domain data are ignored.  The network confidence, however, is independent of the true labels and can thus be directly computed on the target data.

\begin{table*}[t]
	\caption{Comparison of calibration methods for unsupervised domain adaptation (UDA).}
	\label{table:ece_tabx}
     \resizebox{\textwidth}{!}{
	\centering
		\begin{tabular}{l|ccc|ll}
			\toprule 
                Calibration Method &
                Designed for &
                Works without &
                Works on &
                Approach &
                Granularity \\
                &
                domain shift &
                target label &
                target data \\
			\midrule
                Temp. Scaling \cite{Guo2017} & $\times$ & $\times$& $\times$ & -- & Instance level \\
                CPCS\cite{park2020}, TransCal\cite{wang2020} & \checkmark& \checkmark& $\times$ & Importance weight estimation & Instance level\\
                                UTDC (proposed) & \checkmark& \checkmark&\checkmark & Estimates target accuracy & Dataset level \\
			\bottomrule
		\end{tabular}%
      }
	\end{table*}

 We propose a UDA calibration method that computes the confidence and estimates the accuracy directly on the target domain. 
We first assess the accuracy in the target domain. Then we find 
 calibration parameters  that minimize the Expected Calibration Error (ECE)   measure \cite{Naeini2015}  on the target domain. A comparison of typical calibration methods is shown in  \cref{table:ece_tabx}.
Our major contributions include the following:
\begin{itemize}
\item We show that 
 current UDA calibration methods which are all based on the source domain data, rely on an overly  optimistic estimation of the target accuracy. Thus they can't well handle the domain shift problem.
\item We propose a calibration method that is directly  applied to the target domain data, based on a realistic estimation of the accuracy of the adapted model on the target domain.
\end{itemize}
We evaluated our UDA calibration  algorithm on several standard domain  adaptation benchmarks. 
The results on all benchmarks consistently outperformed previous works, thus creating a new standard of calibrating networks for unsupervised domain adaptations.
We show that previously proposed UDA calibration methods don't work at all and thus in this study we propose the first effective method for calibrating a network obtained by an unsupervised domain adaptation.

\section{Background }
\label{sec:background}

Consider a network that classifies an input image  $x$ into $k$ pre-defined categories.
 The network's last layer comprises of  $k$ real numbers $z=(z_1,...,z_k)$ known as \emph{logits}. Each number is the score for one of the $k$ possible classes. The logits are then converted into a soft decision distribution using a \emph{softmax} layer: $p(y=i|x) = \frac{\exp(z_i)}{\sum_j \exp(z_j)}$ where $x$ is the input image and $y$ is the image class.
Despite having the mathematical form of a distribution, the output of the softmax layer does not necessarily represent the true posterior distribution of the classes, and the network often tends to be  over-confidenct in its predictions \cite{Guo2017,Balaji2017,Hein2019}.
The predicted class is calculated from the output distribution  by $\hat{y} = \arg \max_{i} p(y=i|x)= \arg \max_i z_i $. The network \emph{confidence} for this sample is defined by $\hat{p} = p(y=\hat{y}|x) =  \max_{i} p(y=i|x)$. The network \emph{accuracy}  is defined by the probability that the most probable class
$\hat{y}$ is indeed correct. The network is said to be \emph{calibrated} if the estimated confidence coincides with the actual accuracy. 

 The Expected Calibration Error (ECE) \cite{Naeini2015} is the standard  metric used to measure model calibration. It is defined as the expected absolute difference between the model's accuracy and its confidence.
 In practice, the ECE  is computed on a given  validation set  $(x_1,y_1),...,(x_n,y_n)$.  Denote the  predictions and confidence values of the validation set by $(\hat{y}_1,\hat{p}_1),...,( \hat{y}_n,\hat{p}_n)$.
 To compute the ECE measure we first divide the unit interval $[0,1]$ into $M$ equal size bins $b_1,...,b_M$ and let $B_m=\{t| \hat{p}_t \in b_m\}$ be the set of samples whose confidence values   belong to  bin $b_m$. The network  average accuracy at this bin is defined as \(A_m = \frac{1}{|B_m|} \sum_{t \in B_m} \mathbbm{1} \left(\hat{y}_t = y_t\right) \), where $\mathbbm{1}$ is the indicator function, and $y_t$ and $\hat{y}_t$ are the  ground-truth and  predicted labels for $x_t$. 
 The average confidence at bin $b_m$ is defined as
 \(C_m = \frac{1}{|B_m|} \sum_{t \in B_m} \hat{p}_t \).
If the network is under-confident at  bin $b_m$ then $A_m > C_m$  and  vice-versa.
The ECE is defined as follows: \begin{equation}
\mathrm{ECE} = \sum_{m=1}^{M} \frac{|B_m|}{n} \left| A_m - C_m \right|.
\label{ECEdef}
\end{equation}
 The ECE is based on a uniform bin width. If the model is well-trained, most of the
samples should lie within the highest confidence bins.   
Hence, the low confidence bins should be almost empty and therefore have no influence on the computed value of the ECE. For this reason, we  can consider another metric, Adaptive ECE (adaECE)  where the bin sizes are calculated so as to evenly distribute samples between bins  
\cite{Nguyen2015}:
\begin{equation}
\mathrm{adaECE} = \frac{1}{M} \sum_{m=1}^{M} \left| A_m - C_m \right|
\label{adaECEdef}
\end{equation}
such that each bin contains $1/M$ of the data points with similar confidence values. 

Temperature Scaling (TS), is a standard,  highly effective technique for calibrating  the output distribution of a classification network \cite{Guo2017}. It uses a single parameter $T > 0$ to rescale logit scores before applying the softmax function to compute the class distribution.
 Temperature scaling is expressed as follows: 
\begin{equation}
    p_{{\scriptscriptstyle T }}(y=i|x) = \frac{\exp (z_i / T)}{\sum_{j=1}^k\exp (z_j / T)}, \hspace{0.4cm}   i=1,\dots,k
    \label{classcal}
\end{equation} 
s.t. $z_1,...,z_k$ are the logit values obtained by applying the network to input vector $x$. 
 The optimal temperature $T$ for a trained model can be found by  maximizing the log-likelihood $\sum_t \log  p_{{\scriptscriptstyle T }}(y_t|x_t)$ for the held-out validation dataset. Studies show that finding the optimal $T$ by directly minimizing the ECE/adaECE  measures yields better calibration results \cite{mukhoti2020calibrating}.
 The adaECE measure was found to be  much more robust and effective for calibration than ECE. 
 In this study we used the adaECE for both calibration and evaluation.

\section{Unsupervised Target Domain Calibration}
\label{sec:method}
We first formulate the problem of calibration under distribution shift. Assume a network was trained on the source domain. We are given a labeled source domain validation-set dataset, denoted as $\mathcal{S}= \{(x_s^i, y_s^i)\}_{i=1}^{n_s}$ with $n_s$ samples, and an unlabeled target domain dataset $\mathcal{T}=\{x_t^i\}_{i=1}^{n_t}$ with $n_t$ samples. Adapting the network trained on the source domain to the target domain in an unsupervised manner without access to the labels can be achieved using various methods.
 Here, our goal is to calibrate the confidence of the adapted network prediction on samples from the target domain. 
For the sake of simplification, the adapted network will simply be referred to as the ``network'', the source domain validation set data as the ``source data'', and the unlabeled target domain data as the ``target data''.

\begin{figure}
   \centering
   \includegraphics[scale=0.27]{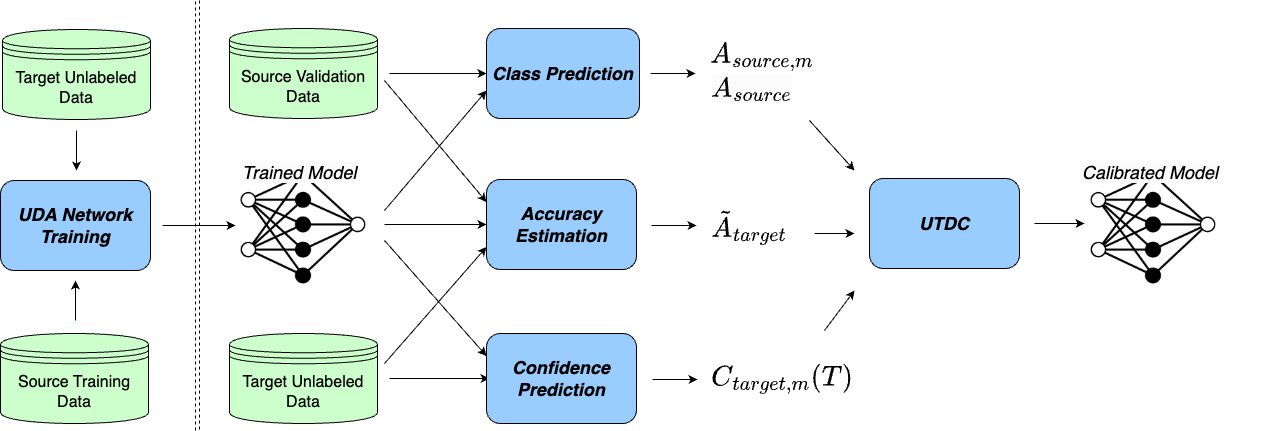}
 \caption{A scheme of the UTDC Calibration Framework. }
 \label{fig:schema}
\end{figure}

Our method involves calibrating the adapted network directly on the target data.
While applying the network on the target domain data allows us to compute its confidence, we cannot determine its accuracy.
Thus, the challenge is to find a reliable estimate of the network accuracy on the target domain.
Our approach is based on the observation that when calibrating by minimizing the adaECE score, we do not need to know whether each individual prediction is correct. Instead, we only need to determine the mean accuracy for each bin.  Fortunately, there are techniques which
 given a trained network, can  estimate the network accuracy 
on  data samples from a new domain without access to their labels  \cite{deng2021labels, Guillory_2021_ICCV, garg2022leveraging, yu2022predicting}.

We next suggest a simple, intuitive, and very effective method that  calibrates the network directly on the target domain.  
We first compute the overall network accuracy  on the source  data $A_{\textrm{source}}$ and  estimate the network accuracy on the target domain (e.g, using \cite{deng2021labels}). Denote the estimated target accuracy by $\tilde{A}_{\textrm{target}}$. 
Next, we divide the source data into $M$ equal-size bins according to their confidence values and compute the corresponding network accuracy $A_{\textrm{source},m}$  at each bin $m$.
We also divide the target data into $M$ equal-size bins according to their confidence values and estimate the binwise accuracy of the target ${A}_{\textrm{target},m}$ by rescaling the binwise accuracy on the source domain  in the following way:
\begin{equation}
\tilde{A}_{\textrm{target},m} =  A_{\textrm{source},m} \cdot \frac{\tilde{A}_{\textrm{target}}}{A_{\textrm{source}}}, \hspace{1cm} m=1,...,M.
\label{targetaccuracy}
\end{equation}
In the next section, we empirically show  that the accuracy ratio between source and target is indeed
similar across the calibration bins.
The estimated network accuracy on the target data 
$\tilde{A}_{\textrm{target}}$ obtained by an unsupervised adaptation is usually lower than its 
accuracy on the source data $A_{\textrm{source}}$. Thus,  this accuracy rescaling provides a more realistic estimation of the bin-wise network average accuracy on the target data. The accuracy ratio 
$\tilde{A}_{\textrm{target}}/A_{\textrm{source}}$ indicates the size of the domain gap or the difficulty of the adaptation task \cite{Zou_2023_ICCV}.

Let $C_{\textrm{target},m}$ be the bin-wise network average confidence values computed on the target data. Substituting the estimated accuracy term, based on the source labeled data (\cref{targetaccuracy}) into the adaECE definition (\cref{adaECEdef}),
yields the following adaECE measure for the target domain in a UDA setup:
\begin{equation}
    \textrm{UDA\mbox{-}adaECE} =  \frac{1}{M} \sum_{m=1}^{M} 
    \left| 
\tilde{A}_{\textrm{target},m}- C_{\textrm{target},m} \right|.
\label{ECEadapt}
\end{equation}

\begin{algorithm*}[t]
\caption { Unsupervised Target Domain Calibration (UTDC) }
\begin{algorithmic}[0]
\State   {\bf input:} A labeled validation set from the source domain, an unlabeled dataset from the target 
domain, and a $k$-class classifier that  was adapted to the target domain.

\vspace{0.1cm}
 \State - Compute the source accuracy $A_{\textrm{source}}$ and 
estimate the target accuracy $\tilde{A}_{\textrm{target}}$ \\ 
\hspace{0.3cm} using a target accuracy estimation technique.

\State - Divide the source points into $M$ equal size sets  based on their  confidence and 
\\ \hspace{0.3cm} compute the binwise mean accuracy: 
$ A_{\textrm{source},m}$, \, $m=1,...,M$.
\State - Estimate the target binwise mean accuracy:
$\tilde{A}_{\textrm{target},m} = A_{\textrm{source},m} \cdot {\tilde{A}_{\textrm{target}}}/{A_{\textrm{source}}} 
 $
\State - Divide the target points into $M$ equal size sets  based  confidence.
\State - For each temperature T compute the target binwize confidence
$C_{\textrm{target},m}(\mbox{T})$ and compute the calibration score:
$$\mathrm{UDA\mbox{-}adaECE}(\mbox{T})
=\sum_{m=1}^{M} \left| 
\tilde{A}_{\textrm{target},m}
- C_{\textrm{target},m}(\mbox{T}) \right|.
$$

\State - Apply a grid search to find the optimal temperature:
  $$\hat{T}= \arg \min_T \mathrm{UDA\mbox{-}adaECE}(T) $$
  \end{algorithmic}
  \label{alg:http}
\end{algorithm*}

For each calibration method whose parameters can be found by minimizing the adaECE measure, we can form a UDA variant in which UDA-adaECE (\cref{ECEadapt}) is minimized instead of adaECE (\cref{adaECEdef}). 
 Examples of these calibration
methods include Temperature Scaling (TS),
 Vector Scaling, Matrix Scaling \cite{Guo2017},  Mix-n-Match \cite{zhang2020mix}, Weight Scaling \cite{frenkel2022miccai}, 
and others.  

We next demonstrate the UDA calibration principle in the case of TS calibration.  
We can determine the temperature that minimizes the UDA-adaECE measure (\cref{ECEadapt}) by conducting a grid search on the possible values. 
 Given the division of the target data into bins, we can  compute the binwise average confidence after temperature calibration by $T$ on the target  ${C}_{\textrm{target},m}(T)$. We can then define the following temperature-dependent adaECE scores: 
\begin{equation}
\mathrm{UDA\mbox{-}adaECE}(T) =  \frac{1}{M}  \sum_{m=1}^{M} \left| 
\tilde{A}_{\textrm{target},m}
- C_{\textrm{target},m}(T) \right|.
\label{ECEt}
\end{equation}
The optimal temperature is thus obtained by applying a grid search to find $T$ that minimizes UDA-adaECE$(T)$.
The proposed  Unsupervised Target Domain Calibration  (UTDC)   algorithm is summarized in \cref{alg:http}. 

{\bf Estimating target accuracy.} A major component of the UTDC method is estimating the target domain accuracy based on unlabeled target domain data.  We next describe several recently suggested estimation algorithms. 
  Deng et al. \cite{deng2021labels} suggested learning a dataset-level regression problem. 
 The first step is to augment the source domain validation set, denoted by $D_s$, using various visual transformations such as resizing, cropping, horizontal and vertical flipping, Gaussian blurring, and others. We then create  $n$ meta-datasets, denoted as $D_1,...,D_{n}$ (in our implementation we set $n=50$). This process preserves the labels so  we can  compute the model's accuracy on these datasets, denoted by $A_1,...,A_{n}$. 
Each dataset $D_i$ is represented as a Gaussian distribution using its mean vector $\mu_i$ and its diagonal covariance matrix $\Sigma_i$. Let $F_i$ be the Fr\'echet  distance \cite{dowson1982frechet}
between the Gaussian representations of $D_s$ and $D_i$. $F_i$ measures the domain gap between
the original dataset $D_s$ and $D_i$.
Next, a linear regression model is fitted to the dataset $(F_1,A_1),...,(F_n,A_n)$ in the form of  $\hat{A}=w\cdot F+b$.
Finally, the linear regression model is employed to predict $\tilde{A}_{\textrm{target}}$, the accuracy of the network on the unlabeled data from the target domain. 
Another method is Average Thresholded Confidence (ATC) \cite{garg2022leveraging} which first selects a threshold $t$ whose error in the source domain matches the expected number of points whose confidence is below $t$. Next, ATC predicts the error
on the target domain which is expressed as the fraction of unlabeled points that obtain a confidence value below that threshold $t$. Let $\hat{p}(x) = \max_i (y=i|x)$ be the network confidence. A  threshold $t$ is calculated to satisfy the equality 
$E_{x \sim \textrm{source}}1_{\{\hat{p}(x) > t\}} =  A_{\textrm{source}}$.
The estimated target accuracy, $\tilde{A}_{\textrm{target}}$, is the expectation $E_{x \sim \textrm{target}}1_{\{\hat{p}(x) > t\}}$.
Finally, the Projection Norm (PN) method \cite{yu2022predicting} uses the model predictions to pseudo-label the test samples and then trains a new model on the pseudo-labels. The discrepancy between the parameters of the new and original models yields the predicted error of the target domain data. %
In Section 5 we compare the UTDC's calibration performance when using each of the target accuracy prediction methods described above.

\section{Experiments}
In this section, we evaluate the capabilities of our UTDC
 technique to calibrate a network on a target domain 
 after applying a UDA procedure. 

{\bf Compared methods.}  
 We compared our method  to six baselines:
(1) Uncalibrated - The adapted classifier as is, without
any post-hoc calibration;
(2-4) Source-TS, Source-VS and source-MS - The adapted network was calibrated by either Temperature Scaling (TS), Vector Scaling (VS) or Matrix scaling (MS) \cite{Guo2017} using the labeled validation set of the source domain; 
(5) CPCS \cite{park2020}, and (6) TransCal \cite{wang2020}, importance weighted UDA calibrators.
We also report Oracle results where TS calibration was applied to the labeled data from the target domain (denoted by Target-TS)  and an Oracle version of our approach (denoted  by UTDC*) where we used the exact accuracy of the adapted model on the target data instead of estimating it.

\begin{table*}[t]
        \caption{AdaECE results  on Office-home (with the lowest in bold) on various UDA classification tasks and models with different calibration methods.}
        \label{table:ece_tab1}
        \centering
       \scalebox{0.8}{ 
                \begin{tabular}{ll|rrrrrrrrr|r}
                        \toprule 
                        {\small UDA}&
               {\small Method}& 
               {\scriptsize $A \rightarrow R$} & 
               {\scriptsize $A \rightarrow C$} &
               {\scriptsize $A \rightarrow  P$} &
               {\scriptsize $C \rightarrow  R$ }&
               {\scriptsize $C \rightarrow  P$} &
               {\scriptsize $C \rightarrow  A$} &
               {\scriptsize $P \rightarrow  R$ }&
               {\scriptsize $P \rightarrow  C$} &
               {\scriptsize $P \rightarrow  A$} &
               Avg \\
                        \midrule

& \small{Uncalibrated} &22.23 & 42.62 & 30.49 & 25.18 & 28.25 & 33.69 & 20.32 & 40.46 & 38.85 & 31.34\\
& \small{Source-TS} &8.09 & 24.43 & 14.89 & 10.00 & 14.17 & 13.85 & 11.14 & 27.42 & 26.60 & 16.73\\
& \small{Source-VS} &10.54 & 27.54 & 19.51 & 12.12 & 14.65 & 15.78 & 11.27 & 31.55 & 27.46 & 18.94\\
& \small{Source-MS} &28.62 & 47.87 & 35.74 & 31.62 & 31.54 & 40.43 & 23.59 & 43.90 & 40.56 & 35.99\\
& \small{CPCS} &15.84 & 49.78 & 23.42 & 14.02 & 16.60 & 18.45 & {\bf 6.31} & 49.21 & 25.62 & 24.36\\
\small{CDAN+E}& \small{TransCal} &6.01 & 27.30 & 9.46 & 16.67 & 16.81 & 21.69 & 19.90 & 41.23 & 39.71 & 22.09\\
 & \small{UTDC} &{\bf 4.46} & {\bf  9.74} & { \bf 7.53} & { \bf 8.36} & { \bf 5.91} & { \bf 8.08} & { 10.45} & { \bf 7.46} & { \bf 9.37} & { \bf 7.93}\\
\cline{2-12} 
& \small{UTDC*} &4.30 & 5.93 & 7.41 & 7.85 & 4.62 & 10.16 & 10.76 & 4.55 & 9.54 & 7.24\\
& \small{Target-TS} &3.97 & 5.05 & 7.19 & 4.07 & 4.39 & 7.07 & 2.32 & 4.39 & 8.57 & 5.22\\
 \hline & \small{Uncalibrated} &19.90 & 39.19 & 26.75 & 24.47 & 26.33 & 33.53 & 20.25 & 40.06 & 39.25 & 29.97\\
& \small{Source-TS} &6.90 & 19.80 & 7.93 & 6.54 & 7.01 & 16.01 & 15.68 & 27.87 & 30.97 & 15.41\\
& \small{Source-VS} &10.15 & 25.83 & 15.31 & 12.13 & 10.70 & 17.90 & {\bf 14.69} & 32.40 & 31.64 & 18.97\\
& \small{Source-MS} &30.78 & 52.03 & 38.39 & 35.44 & 35.45 & 44.21 & 26.40 & 45.87 & 43.33 & 39.10\\
& \small{CPCS} &13.90 & 50.16 & 21.32 & {\bf 3.62} & 7.25 & 34.74 & 25.86 & 22.66 & 27.97 & 23.05\\
\small{DANN+E}& \small{TransCal} &7.21 & 27.42 & 12.36 & 17.81 & 15.43 & 29.93 & 24.64 & 46.61 & 45.83 & 25.25\\
  & \small{UTDC} &{\bf 4.14} & {\bf 5.86} & {\bf 5.47} & { 10.28} & {\bf 3.89} & { \bf 6.67} & { 15.33} & {\bf  5.70} & { \bf 12.65} & {\bf 7.78}\\
\cline{2-12}
& \small{UTDC*} &2.68 & 4.70 & 4.37 & 8.55 & 4.00 & 4.53 & 14.60 & 3.97 & 6.16 & 5.95\\
& \small{Target-TS} &2.68 & 2.76 & 3.67 & 2.24 & 3.16 & 2.99 & 1.15 & 1.62 & 4.55 & 2.76\\
 \hline & \small{Uncalibrated} &16.82 & 31.28 & 23.11 & 17.22 & 20.46 & 27.38 & 15.88 & 33.81 & 30.13 & 24.01\\
& \small{Source-TS} &6.33 & 16.41 & 13.22 & {\bf 2.83} & {\bf 5.00} & 15.82 & {\bf 10.91} & 29.09 & 23.61 & 13.69\\
& \small{Source-VS} &10.03 & 25.58 & 15.86 & {8.10} & 8.23 & 15.18 & 11.86 & 33.08 & 27.24 & 17.24\\
& \small{Source-MS} &31.61 & 50.68 & 41.31 & 34.23 & 36.48 & 44.23 & 25.49 & 44.75 & 40.17 & 38.77\\
& \small{CPCS} &8.89 & 33.56 & 19.99 & 25.29 & 9.62 & {\bf 12.82} & 16.87 & 27.49 & 45.93 & 22.27\\
\small{DANN}& \small{TransCal} &7.63 & 29.15 & 22.20 & 22.64 & 22.97 & 37.66 & 26.11 & 50.85 & 47.53 & 29.64\\
& \small{UTDC} & { \bf  5.15} & {\bf 4.87} & {\bf 11.24} & { 8.63} & {  5.23} & { 15.08} & { 18.62} & {\bf  12.62} & { \bf 11.23} & \textbf{ 10.30}\\
\cline{2-12} 
& \small{UTDC*} &2.80 & 5.49 & 6.21 & 6.20 & 3.38 & 3.44 & 12.61 & 5.00 & 4.67 & 5.53\\
& \small{Target-TS} &2.45 & 2.38 & 4.65 & 2.08 & 1.73 & 2.16 & 1.22 & 2.35 & 2.92 & 2.44\\

            \bottomrule
        \end{tabular}%
            }
        \end{table*}
{\bf Datasets.} We report experiments on four standrad real-world domain adaptation benchmarks, Office-home \cite{venkateswara2017deep}, Office-31 \cite{saenko2010adapting}, VisDa-2017 \cite{peng2017visda}, and DomainNet \cite{peng2019moment}. Office-home includes four domains - Art, Real-World, Clipart and Product, represented as A, R, C, and P in the experiments.  Office-31 contains three domains -  Amazon, Webcam and  DSLR, denoted  A, W, and D. VisDa-2017 is a simulation-to-real dataset for domain adaptation with over 280,000 images across 12 categories. DomainNet has six domains - Clipart, Infograph, Painting, Quickdraw, Real and Sketch, denoted  C, I, P, Q, R, and S.

\begin{table*}[h!]
        \caption{AdaECE results on Office-31  (with the lowest in bold) on various UDA  classification tasks and models with different calibration methods.}
        \label{table:ece_tab2}
        \centering
         \scalebox{0.8}{ 
               \begin{tabular}{ll|rrrrrr|r}
                        \toprule 
                        {\small UDA Method}&
               {\small Method}& 
               {\scriptsize $A \!\rightarrow\! W$} & 
               {\scriptsize $A \!\rightarrow\! D$} &
               {\scriptsize $W \!\rightarrow\!  A$} &
               {\scriptsize $W \!\rightarrow\!  D$ }&
               {\scriptsize $D \!\rightarrow\!  A$} &
               {\scriptsize $D \!\rightarrow\!  W$} &
               Avg \\
                        \midrule

& \small{Uncalibrated} &11.5 & 10.53 & 29.63 & 1.21 & 29.08 & {\bf 1.33} & 13.88\\
& \small{Source-TS} &6.03 & 7.43 & 33.21 & {\bf 0.86} & 27.25 & 2.12 & 12.82\\
& \small{Source-VS} &{\bf 3.74} & 7.10 & 33.75 & 1.52 & 32.98 & 1.42 & 13.42\\
& \small{Source-MS} &12.15 & 16.72 & 30.76 & { 1.02} & 29.99 & 1.38 & 15.34\\
& \small{CPCS} &9.67 & 12.66 & 33.47 & 1.11 & 28.16 & 2.18 & 14.54\\
\small{CDAN+E}& \small{TransCal} &3.78 & 9.45 & 34.43 & 1.27 & 33.68 & 1.56 & 14.03\\

& \small{UTDC} &4.19 & {\bf 5.18} & {\bf 5.15} & 1.20 & {\bf 5.14} & 2.18 & {\bf 3.84}\\
\cline{2-9} 
& \small{UTDC*} &3.82 & 5.18 & 5.09 & 1.13 & 5.36 & 2.18 & 7.13\\
& \small{Target-TS} &3.44 & 4.67 & 3.32 & 0.75 & 3.20 & 0.89 & 2.71\\
 \hline & \small{Uncalibrated} &13.05 & 13.55 & 28.29 & 0.87 & 27.15 & 1.68 & 14.10\\
& \small{Source-TS} &5.18 & 9.29 & 26.93 & 1.31 & 26.44 & 2.44 & 11.93\\
& \small{Source-VS} & {\bf 4.63} & 8.24 & 36.64 & {\bf 0.87} & 31.35 & 1.55 & 13.88\\
& \small{Source-MS} &18.01 & 14.02 & 31.10 & 1.09 & 28.51 & 1.51 & 15.71\\
& \small{CPCS} &15.58 & 6.81 & 33.97 & 1.99 & 32.69 & {\bf 1.14} & 15.36\\
\small{DANN+E}& \small{TransCal} &7.98 & 5.63 & 34.53 & 1.57 & 31.12 & 1.59 & 13.74\\
& \small{UTDC} &5.25 & {\bf 5.33} & {\bf 8.99} & 1.40 & {\bf 12.26} & 2.41 & {\bf 5.94}\\
\cline{2-9} 
& \small{UTDC*} &4.87 & 6.10 & 6.86 & 1.40 & 6.53 & 2.44 & 4.70\\
& \small{Target-TS} &3.98 & 4.77 & 2.87 & 0.85 & 2.80 & 0.82 & 2.68\\
 \hline & \small{Uncalibrated} &10.66 & 12.59 & 23.03 & 1.77 & 24.43 & 2.93 & 12.57\\
& \small{Source-TS} &3.89 & {\bf 7.17} & 29.58 & {\bf 0.98} & 30.71 & 4.43 & 12.79\\
& \small{Source-VS} &3.88 & 7.64 & 34.50 & 1.44 & 32.31 & 2.84 & 13.77\\
& \small{Source-MS} &21.06 & 24.70 & 28.81 & 1.35 & 28.45 & {\bf 1.30} & 17.61\\
& \small{CPCS} &16.96 & 10.10 & 33.69 & 2.61 & 35.39 & 4.80 & 17.26\\
\small{DANN}& \small{TransCal} &10.36 & 15.62 & 87.02 & 2.31 & 45.79 & 6.00 & 27.85\\
& \small{UTDC} &{\bf 3.71} & 8.70 & {\bf 5.14} & 2.61 & {\bf 9.26} & 5.23 & {\bf 5.78}\\
\cline{2-9} 
& \small{UTDC*} &5.04 & 7.52 & 5.54 & 2.61 & 12.25 & 6.54 & 6.58\\
& \small{Target-TS} &3.53 & 4.12 & 2.79 & 0.97 & 3.19 & 1.94 & 2.76\\

            \bottomrule
        \end{tabular}%
       }
        \end{table*}

{\bf Implementation details.}
We followed the experiment setup described in  \cite{wang2020} and used their code to implement CPCS and TransCal baselines.
Following \cite{wang2020}, we implemented three  different UDA techniques; namely, DANN \cite{ganin2016domain}, DANN+E and CDAN+E \cite{long2018conditional}. 
The performance of more recent UDA models (e.g. \cite {liang2021domain, jin2020minimum, cui2020towards}) on the target domain of the evaluated datasets is slightly better  but  is still much worse than the performance on the source domain. 
In most experiments we used the Meta target domain accuracy estimation  \cite{deng2021labels} unless stated otherwise.
We provide a code implementation of our method for 
reproducibility\footnote{\href{https://github.com/cobypenso/unsupervised-target-domain-calibration}{https://github.com/cobypenso/unsupervised-target-domain-calibration}}.

https://github.com/cobypenso/unsupervised-target-domain-calibration

\begin{table}[h!]
        \caption{adaECE results on VisDA Task $S \rightarrow R$, for various calibration methods.}
        \vspace{0.1cm}
	\label{table:adaece_visda}
   \centering
   \scalebox{0.83}{ 
               \setlength\tabcolsep{7pt}
                \begin{tabular}{l|rrr|r}
                        \toprule 
               {\small Method}& 
               {\footnotesize DANN} & 
               {\footnotesize DANN+E} &
               {\footnotesize CDAN+E} &
               {\footnotesize Avg} \\
                        \midrule
    \small{Uncalibrated} &33.23 & 31.79 & 29.88& 31.63\\		
    \small{Source-TS} &26.54 & 18.66  & 23.38& 34.29 \\		
    \small{Source-VS} &38.22 & 36.96 & 28.48& 34.55\\		
    \small{Source-MS} &41.19 & 38.17 & 30.87& 36.74\\		
    \small{CPCS} &31.86 & 11.08 & 26.88& 23.27\\		
    \small{TransCal} &43.52 & 35.93 & 36.71& 38.72\\		
    \small{UTDC} &\textbf{13.07} & \textbf{6.61}  & \textbf{3.85}& \textbf{7.84} \\	
    \cline{1-5}		
    \small{UTDC*} & 2.31 & 1.94 & 2.57& 2.27\\		
    \small{Target-TS} &2.02 & 1.84 & 2.21 & 2.02\\		
\bottomrule
    \end{tabular}%
 }
\end{table}

\begin{table}[h!]
        \caption{adaECE results on DomainNet for various UDA  classification tasks and models with different calibration methods.}
        \label{table:ece_tab4}
          \centering
         \scalebox{0.82}{ 
                      \begin{tabular}{ll|rrrrrr|r}
                        \toprule 
                        {\small UDA}&
               {\small Method}& 
               {\scriptsize $S \!\!\rightarrow\!\! R$} & 
               {\scriptsize $S \!\!\rightarrow\!\! P$} &
               {\scriptsize $P \!\!\rightarrow\!\!  R$} &
               {\scriptsize $P \!\!\rightarrow\!\!  S$ }&
               {\scriptsize $R \!\!\rightarrow\!\!  S$} &
               {\scriptsize $R \!\!\rightarrow\!\!  P$} &
               Avg \\
                        \midrule

& \small{Uncalibrated} &14.65 & 18.70 & 18.06 & 22.98 & 19.13 & 13.77 & 17.88\\
& \small{Source-TS} &12.68 & 14.48 & 11.51 & 12.76 & 13.56 & 9.60 & 12.39\\
& \small{Source-VS} &10.70 & 9.56 & 11.49 & 14.94 & 13.35 & 9.31 & 11.56\\
& \small{Source-MS} &22.24 & 25.28 & 23.43 & 30.93 & 22.55 & 18.07 & 23.75\\
\small{CDAN+E}& \small{CPCS} &9.41 & 11.20 & 13.26 & 17.06 & 17.16 & 11.86 & 13.32\\
& \small{TransCal} &12.50 & 20.82 & 16.41 & 28.85 & 36.70 & 28.23 & 23.92\\
& \small{UTDC} & {\bf 6.06} & {\bf 5.17} & {\bf 6.48} & {\bf 4.75} & {\bf 8.85} & {\bf 8.32} & \textbf{6.61}\\
\cline{2-9} 
& \small{UTDC*} &5.07 & 6.78 & 4.86 & 3.56 & 5.19 & 6.86 & 5.38\\
& \small{Target-TS} &1.31 & 1.35 & 2.18 & 1.39 & 1.25 & 1.07 & 1.42\\
 \hline & \small{Uncalibrated} &15.03 & 17.77 & 17.57 & 24.54 & 21.08 & 16.63 & 18.77\\
& \small{Source-TS} &10.12 & 12.20 & 10.31 & 11.75 & 11.76 & 10.69 & 11.14\\
& \small{Source-VS} &9.71 & 14.25 & 11.85 & 19.42 & 16.88 & 12.15 & 14.04\\
& \small{Source-MS} &23.68 & 28.77 & 24.18 & 35.03 & 24.94 & 20.91 & 26.25\\
\small{DANN+E}& \small{CPCS} &13.20 & 6.41 & 12.51 & 12.81 & {\bf  7.73} & {\bf 10.95} & 10.60\\
& \small{TransCal} &14.56 & 19.85 & 16.14 & 29.19 & 34.98 & 28.96 & 23.95\\
& \small{UTDC} & {\bf 6.39} & {\bf 6.07}  & {\bf 6.54} & {\bf 6.84} & { 11.24} & { 11.94} & \textbf{8.17}\\
\cline{2-9} 
& \small{UTDC*} &3.97 & 5.72 & 5.23 & 6.64 & 6.73 & 8.32 & 6.10\\
& \small{Target-TS} &1.24 & 1.19 & 1.60 & 1.03 & 1.10 & 0.84 & 1.17\\
 \hline & \small{Uncalibrated} &10.98 & 13.52 & 12.65 & 18.04 & 15.42 & 10.96 & 13.59\\
& \small{Source-TS} &7.33 & 8.63 & 9.50 & 10.11 & 10.99 & 9.15 & 9.29\\
& \small{Source-VS} &8.92 & 14.43 & 11.21 & 16.90 & 15.86 & 10.86 & 13.03\\
& \small{Source-MS} &22.51 & 27.48 & 21.97 & 31.46 & 24.53 & 19.72 & 24.61\\
\small{DANN}& \small{CPCS} &7.02 & 7.37 & 14.60 & 15.83 & 15.42 & 8.88 & 11.52\\
& \small{TransCal} &14.83 & 22.09 & 16.38 & 30.37 & 37.84 & 29.92 & 25.24\\
& \small{UTDC} & {\bf 5.82}  & {\bf 5.84} & {\bf 6.30} & {\bf 9.24} & {\bf 5.80} & {\bf 7.53} & \textbf{6.76}\\
\cline{2-9} 
& \small{UTDC*} &4.34 & 5.46 & 4.71 & 7.34 & 6.53 & 6.81 & 5.87\\
& \small{Target-TS} &1.07 & 1.25 & 1.06 & 0.90 & 1.53 & 1.61 & 1.24\\

            \bottomrule
        \end{tabular}%
            }
        \end{table}

{\bf Calibration results.}
   \cref{table:ece_tab1}, \cref{table:ece_tab2}, \cref{table:adaece_visda} and \cref{table:ece_tab4} report the calibration results (computed by adaECE with 15 bins) on  Office-home, Office-31, VisDA, and DomainNet respectively.
 The results show that UTDC achieved significantly better results than the baseline methods on all tasks.  The calibration obtained by previous  IW-based methods was slightly better (but in some cases even worse) than a network with no calibration or a network that was calibrated on the source domain. 
 In contrast, the adaECE score obtained by UTDC was almost as good as the adaECE obtained by an oracle that had access to the labels of the domain samples. 
In addition to the adaECE evaluation measure,   \cref{table:bs_log_tab1} reports the average calibration results over all Office-home tasks, using three other calibration metrics:   ECE, Negative Log-Likelihood (NLL) and Brier Score (BS) \cite{brier1950}. The same trends as above were observed. 

 \begin{table}[h!]
        \caption{Calibration metrics results of various UDA calibration methods on the Office-home tasks.}
        \vspace{0.1cm}
	\label{table:bs_log_tab1}
                     \centering
         \scalebox{0.82}{ 
\begin{tabular}{lrrrrrrrrr}
\toprule
   & \multicolumn{3}{c}{CDAN+E}  & \multicolumn{3}{c}{DANN+E}  & \multicolumn{3}{c}{DANN}\\ 
       method &  BS &   NLL &    ECE &  BS &   NLL &    ECE &  BS &   NLL &    ECE\\
\midrule
  Uncalibrated &   0.74 &  3.40 &  31.32  &   0.76 &  3.07 &  29.92 &   0.75 &  2.75 &  24.08\\
        Source-TS &   0.65 &  2.18 &  16.79 &   0.67 &  2.21 &  15.40 &   0.71 &  2.37 &  13.71 \\
                 CPCS &   0.71 &  3.48 &  24.46 &   0.72 &  3.08 &  23.12 &   0.76 &  2.87 &  22.37 \\
          TransCal &   0.69 &  2.70 &  22.12 &   0.73 &  3.08 &  25.22 &   0.81 &  3.72 &  29.71\\
              UTDC &  {\bf 0.62} & {\bf 1.95} & {\bf  8.01} & {\bf 0.64} &  {\bf2.01} &  {\bf 7.81}
           &   {\bf 0.69} & {\bf 2.26} & {\bf 10.35}            \\
      \midrule
                  UTDC* &   0.62 &  1.95 &   7.21 &   0.63 &  1.99 &   5.94 &   0.68 &  2.18 &   5.53\\
                   Target-TS &   0.61 &  1.92 &   5.41 &   0.63 &  1.96 &   2.72 &   0.68 &  2.14 &   2.78\\
\bottomrule
\end{tabular}
    }
\end{table}

\begin{table}[h]
        \caption{Computed temperature on various UDA Office-home tasks, and calibration methods using CDAN+E.}
	\label{table:t_tab1}
        \centering
                     \scalebox{0.82}{ 
                \begin{tabular}{ll|ccccccccc|r}
                        \toprule 
                        {\small UDA}&
               {\small Method}& 
               {\scriptsize $A \!\!\rightarrow\!\! R$} & 
               {\scriptsize $A \!\!\rightarrow\!\! C$} &
               {\scriptsize $A \!\!\rightarrow\!\!  P$} &
               {\scriptsize $C \!\!\rightarrow\!\!  R$ }&
               {\scriptsize $C \!\!\rightarrow\!\!  P$} &
               {\scriptsize $C \!\!\rightarrow\!\!  A$} &
               {\scriptsize $P \!\!\rightarrow\!\!  R$ }&
               {\scriptsize $P \!\!\rightarrow\!\!  C$} &
               {\scriptsize $P \!\!\rightarrow\!\!  A$} &
               Avg \\
                        \midrule

& \small{Source-TS} &1.96 & 2.02 & 2.02 & 1.87 & 1.90 & 2.06 & 1.63 & 1.72 & 1.68 & 1.87\\
& \small{CPCS} &1.46 & 0.57 & 1.49 & 1.68 & 1.75 & 2.05 & 1.93 & 0.50 & 1.73 & 1.46\\
& \small{TransCal} &2.12 & 1.86 & 2.39 & 1.50 & 1.74 & 1.62 & 1.03 & 0.96 & 0.95 & 1.57\\
   \small{CDAN+E}& \small{UTDC} &2.27 & 2.90 & 2.91 & 1.97 & 2.44 & 2.54 & 1.67 & 2.93 & 2.89 & 2.50\\
\cline{2-12}
& \small{UTDC*}  &2.29 & 3.21 & 2.68 & 2.00 & 2.62 & 2.30 & 1.65 & 3.41 & 2.90 & 2.56\\
& \small{Target-TS} &2.36 & 3.61 & 2.73 & 2.42 & 2.73 & 2.81 & 2.24 & 3.49 & 3.37 & 2.86\\

            \bottomrule
                \end{tabular}%
           }
        \end{table}

\begin{figure}[ht!]
         \resizebox{\textwidth}{!}{
    \includegraphics[scale=0.34]{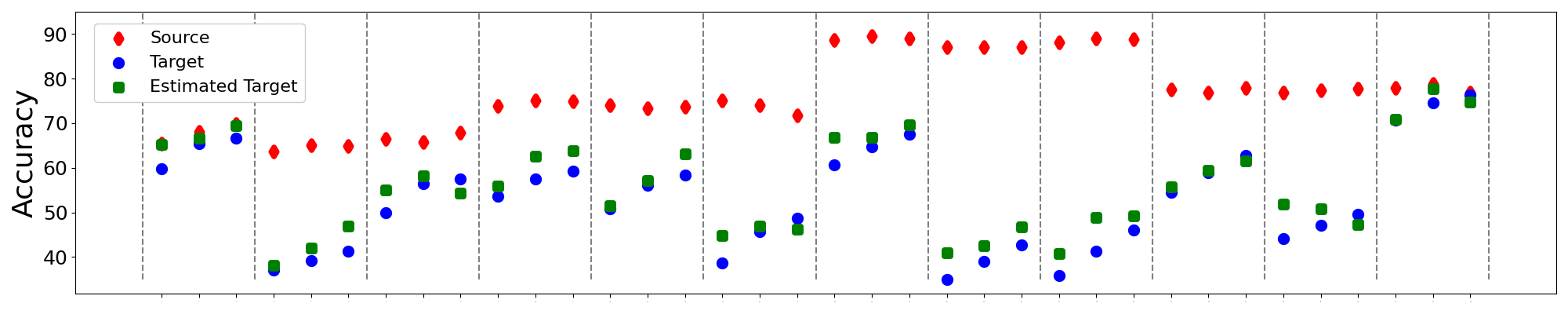}
     }
        { \noindent \hspace{1.6cm}  
         \resizebox{\textwidth}{!}
        { \hspace{2cm}
        $A \rightarrow R$ \hspace{0.29cm}
    { $A \rightarrow C$\hspace{0.29cm}
    $A \rightarrow P$\hspace{0.29cm}
    $C \rightarrow R$\hspace{0.29cm}
    $ C \rightarrow P$\hspace{0.29cm}
    $ C \rightarrow A$\hspace{0.29cm}
    $ P \rightarrow R$ \hspace{0.29cm}
    $ P \rightarrow C$\hspace{0.29cm}
    $ P \rightarrow A$\hspace{0.29cm}
    $ R \rightarrow A$\hspace{0.29cm}
    $ R \rightarrow C$ \hspace{0.29cm}
    $ R \rightarrow P$ \hspace{1.29cm} 
    }}}
    \caption{Average accuracy on Office-home tasks for 
    the three UDA techniques (DANN, DANN+E, CDAN+E).}
    \label{fig:my_label}
\end{figure}

 \section{Analysis}
 We next illustrate and analyze several key features of the proposed method. 
 
{\bf Accuracy gap between source and target.}
\label{subsec:understand_gap}
To gain a better understanding of the reasons 
why our method performs better than IW based methods, we first discuss the accuracy of the adapted models on the source and target domains. \cref{fig:my_label} presents the accuracy on the source and target domains for three UDA techniques. It shows that even after adaptation to the target,  the model's performance on the source samples is consistently better than  its performance on the target samples, especially in cases of large  domain gaps.
Hence, using the network accuracy on the source to  estimate  the network's accuracy on the target while minimizing the ECE measure is misleading  
because the over-optimistic accuracy estimation leads to 
 a scaling  temperature that is too small.
 \cref{table:t_tab1} compares the optimal temperatures computed by the  calibration methods. In all the baseline methods the computed calibration temperature was lower than the optimal value. This results in  poorer calibration performance, as seen in  \cref{table:ece_tab1}, \cref{table:ece_tab2}, \cref{table:adaece_visda}, and \cref{table:ece_tab4}. By contrast,  the temperature computed by all the  UTDC variants was much closer to the  optimal temperature computed by the Oracle method that had access to the target labels. 
 \cref{fig:my_label} also presents the estimated accuracy of the adapted model on the target domain. This estimation is close to the true accuracy. Thus, when it is combined with the confidence computed on the target domain, we  obtain a calibrated mode.

\begin{figure*}[t!]
 \resizebox{\textwidth}{!}{
    \centering
    \includegraphics[trim= 180 00 180 0, clip, scale=0.30]{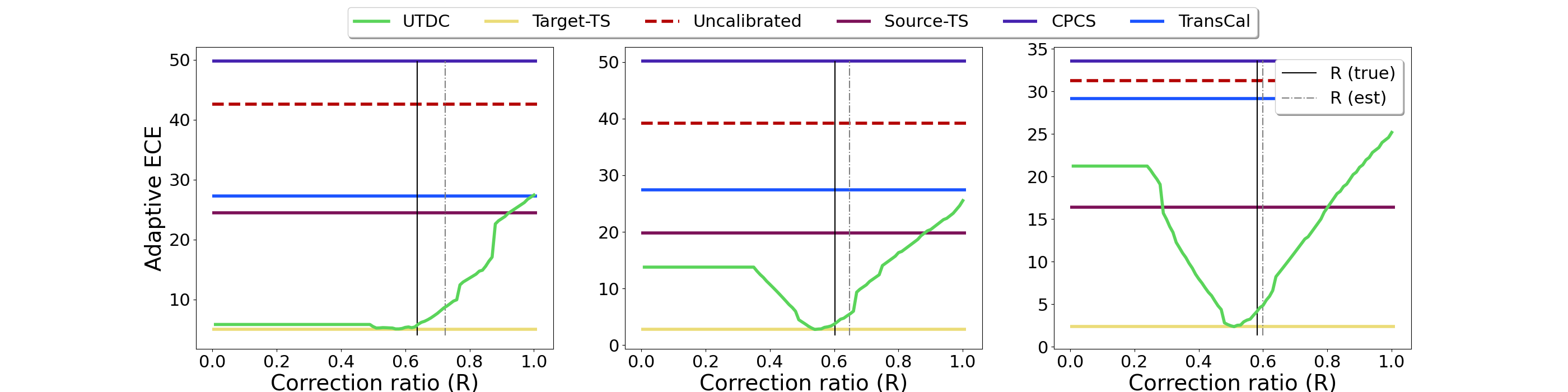} }
          \vspace{0.2cm}
    \hspace{1.2 cm} 
    (a) CDAN+E \hspace{2.0cm} 
    (b) DANN+E \hspace{2.2cm} 
    (c) DANN
      \caption{adaECE results as a function of the correction ratio $R$ on Office-Home, $A \rightarrow C$ task. }
    \label{fig:correction_ratio}
         \end{figure*}

     \begin{figure}[h!]
    \centering
    \includegraphics[scale=0.14]{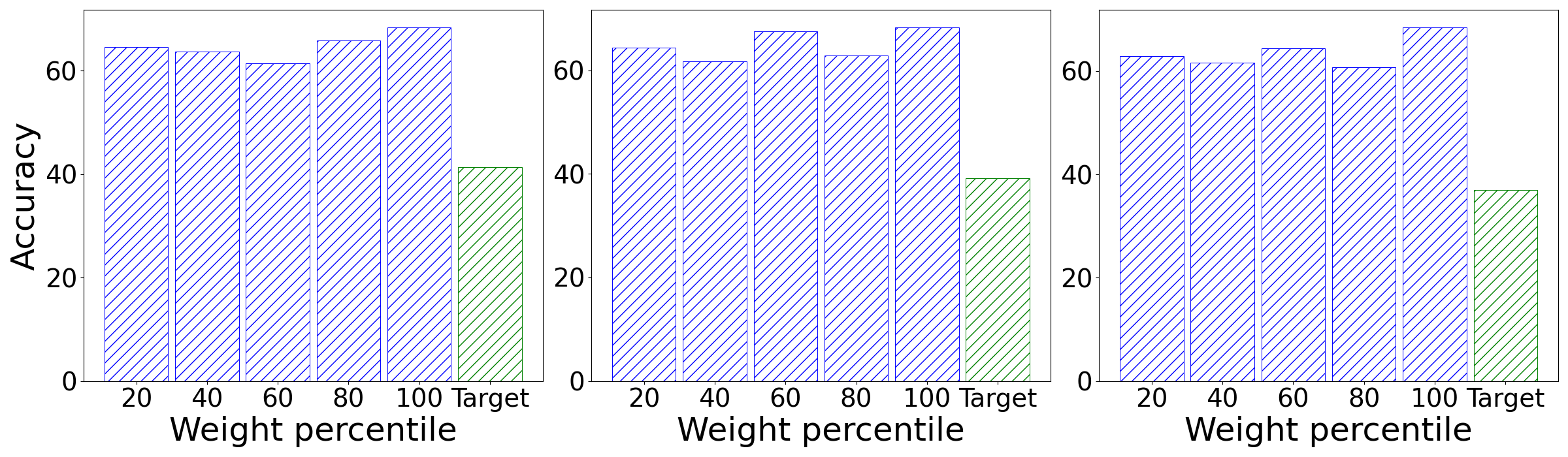} \\
       \hspace{0.65cm}
    (a) CDAN+E \hspace{.6cm}  (b) DANN+E \hspace{.6cm} (c) DANN \hspace{1.4cm} 
    \caption{Accuracy of $k$-th percentile source images based on their probability of being classified as target \cite{wang2020}, compared to target accuracy (Office-home, $A \rightarrow C$). }
    \label{fig:source_acc_gap}
\end{figure}
\begin{figure}[h!!]
    \centering
    \includegraphics[scale=0.15]{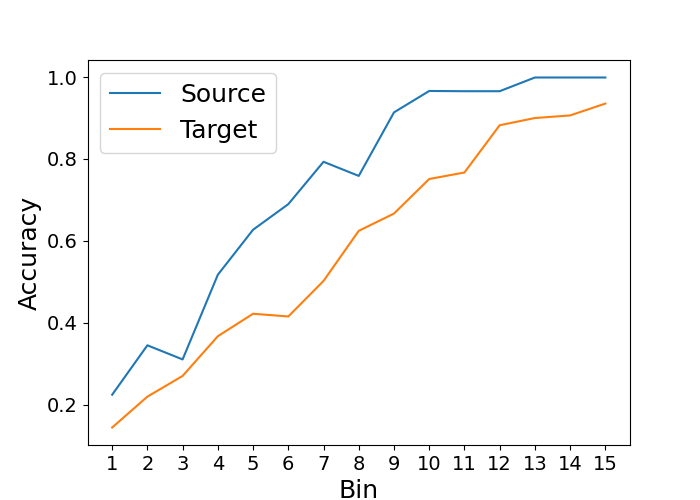}
    \includegraphics[scale=0.15]{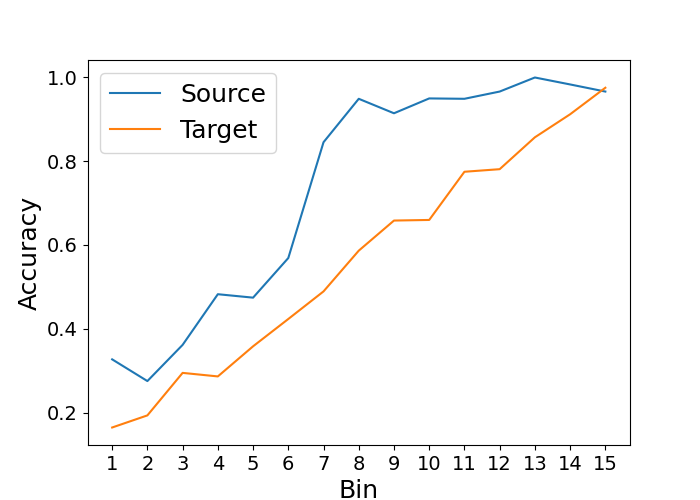}
    \includegraphics[scale=0.15]{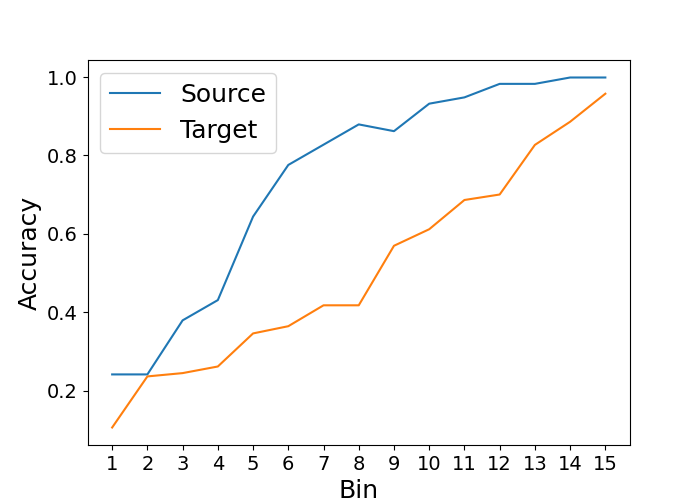}  \\
     (a) CDAN+E \hspace{0.6cm}  (b) DANN+E \hspace{0.6cm} (c) DANN 
    \caption{Accuracy per bin for source and target images. The results are shown on the Office-home $C \rightarrow P$ task. }
    \label{fig:acc_ratio_bins}
\end{figure}
{\bf Sensitivity of UTDC to the target accuracy prediction.} 
\label{sec:sweep}
UTDC is based on estimating the binwise average network accuracy on the target domain data from the labeled source domain data. This estimation is done by computing the ratio ${\tilde{A}_{\textrm{target}}}/{A_{\textrm{source}}}$ between the estimated target accuracy  and the source accuracy. We next analyze the sensitivity of our calibration method to errors in estimating $A_{\textrm{target}}$.
Let  $ R(\textrm{true}) = {A_{\textrm{target}}}/{A_{\textrm{source}}}$ and $ R(\textrm{estimated}) = {\tilde{A}_{\textrm{target}}}/{A_{\textrm{source}}}$
be the true and estimated ratio used by UTDC* and UTDC respectively.
In principle, any number  $0\!<\!R$ can be used to obtain an  estimation of the binwise target accuracy:  $\tilde{A}_{\textrm{target},m} = {A}_{\textrm{source},m} \cdot R$. We can thus find the temperature that minimizes the
adaECE function on the target data as a function of $R$: 
$\hat{T}(R) = \arg \min_T \mathrm{adaECE}_R(T)$
where  
$$ \mathrm{adaECE}_R(T) =
 \frac{1}{M} \sum_{m=1}^{M}\left| 
{A}_{\textrm{source},m} \cdot R - C_{\textrm{target},m}(T) \right|.
$$
\cref{fig:correction_ratio} shows the adaECE measure on the target data after temperature scaling by $\hat{T}(R)$  as a function of the ratio $R$ for the task Office-home $A \rightarrow C$.  It shows  that with the  appropriate choice of $R$  we can achieve the calibration level of the Oracle TS-target algorithm (the case where target labels are known). This means that the difference in accuracy is indeed the main reason for the calibration degradation caused by methods that try to calibrate the target domain using the source data.    
Specifically, as the ratio $R$  drops towards $R$(true), the adaECE improves and approaches the  Oracle TS-target calibration. In addition, the adaECE reaches a minimum near $R$(true) and $R$(estimated).  Finally, there is a range of correction ratios where UTDC is better by a large margin than other baselines, thus providing a tolerance for error and resilience in estimating $\tilde{A}_{\textrm{target}}$.

{\bf The problem with the IW assumption.}
We showed that our method achieves better results by explicitly addressing the accuracy gap between the source and  target domains caused by the domain shift.
Previous methods based on importance weights 
 \cite{park2020,wang2020} rely on re-weighting the source data based on their proximity to the target data, i.e., concentrating on source samples that resemble the target and attributing less weight to others. We computed the target similarity weights associated with each sample in the source validation set and divided them into $20\%$ percentile subsets.  \cref{fig:source_acc_gap}  shows the average accuracy of each group and the average target accuracy.  
It shows that the source accuracy is similar in all bins regardless of the similarity to the target. Thus the IW assumption
that source samples that are classified as targets are more relevant for calibrating the target prediction is wrong.

\begin{table}[t]
\centering
\caption{AdaECE results for variations of UTDC based on different methods of domain accuracy estimation.}
\label{table:diff_est_methods__ece}
      \centering
         \scalebox{0.83}{ 
    \begin{tabular}{l|rrrr}
        \toprule 
        {\small Method}&
           {\small Office-home}& 
           {\small Office-31} & 
           {\small VisDA} &
           {\small DomainNet} \\
        \midrule
            \small{Uncalibrated} & 28.44 & 13.51 &  31.63 & 16.74\\
            \small{UTDC-Meta}\cite{deng2021labels} & 8.67 & 6.96 &  7.84 & 7.18\\
            \small{UTDC-ATC}\cite{garg2022leveraging} & 10.12  & 7.47 &  5.68 & 8.01\\
            \small{UTDC-PN}  \cite{yu2022predicting} &  11.55 &  7.83 &  10.20 & 8.63 \\
            \small{UTDC*} & 6.24  & 6.13  & 2.27 & 5.78\\
        \bottomrule
    \end{tabular}%
    }
\end{table}
\begin{table}[t]
\centering
\caption{Comparison of several target domain accuracy estimation methods measured by   $|ACC(True) - ACC(Est)|$.}
\label{table:diff_est_methods__acc}
      \centering
         \scalebox{0.83}{ 
    \begin{tabular}{l|cccc}
        \toprule 
        {\small Method}&
           {\small Office-home}& 
           {\small Office-31} & 
           {\small VisDA} &
           {\small DomainNet}\\
        \midrule 
            \small{Meta}\cite{garg2022leveraging} & 3.31 & 2.81 &  4.96 & 3.10\\
            \small{ATC}\cite{garg2022leveraging} & 5.05  & 3.37 &  3.48 & 4.25\\
            \small{PN} \cite{yu2022predicting} &  6.26 & 4.85 & 6.30 & 5.91 \\
        \bottomrule
    \end{tabular}%
    }
\end{table}

{\bf Accuracy ratio across bins.}
Our method computes $\tilde{A}_{\textrm{target},m}$ by re-scaling $A_{\textrm{source},m}$ with the same ratio for all bins, as defined in  \cref{targetaccuracy}. This estimation is based on the assumption that the accuracy ratio between the source and the target is similar across the bins. To illustrate the validity of this assumption, \cref{fig:acc_ratio_bins} shows the accuracy of the adapted network  at  each bin, for the source and target data.

{\bf Different target accuracy estimation methods.} 
\label{sec:other_est}
Our UTDC method requires an estimation step of the target domain accuracy without labels. In all the experiments  reported above we used the Meta method \cite{deng2021labels}. 
We next examine combining UTDC with two other methods for target domain accuracy estimation:   ATC \cite{garg2022leveraging} and PN \cite{yu2022predicting}. 
  We implemented 3 variations of UTDC, dubbed  UTDC-Meta, UTDC-ATC, and UDTC-PN based on the estimated target accuracy that was used.  We also report results for UTDC*  based on the true target accuracy. 
   \cref{table:diff_est_methods__ece} and \cref{table:diff_est_methods__acc} present the average calibration results and the discrepancy between the estimated and actual accuracy, respectively.
  The results indicate that UTDC achieved the best calibration performance out of all the three target accuracy estimation methods examined, thus reinforcing the observed low sensitivity of UTDC to the precision of target accuracy predictions. This underscores the compatibility of UTDC with existing methods for network calibration under unsupervised domain shift.
We also found that using UTDC-Meta yields better results, while UTDC-ATC exhibits improved performance and ease of implementation, since the ATC method is much simpler to implement and requires a small computational effort.

\section{Conclusion}

This work considered the problem of network calibration in an unsupervised domain adaptation setup. We first showed that the main problem with calibration using the labeled data from the source domain is the accuracy difference between the domains. We then showed that methods that are based on importance weighting do not address this problem, which causes them to fail.  Our key idea with respect to previous methods 
is replacing the over-optimistic accuracy estimation, performed on the labeled data from the source domain, with the actual accuracy of the adapted model on the target domain, and calibrating directly over the target examples.    We compared this solution to previous methods and showed that it consistently and significantly improved the calibration results on the target domain.
We concentrated here on parametric calibration methods of classification tasks under domain shift. Possible future research directions include applying similar strategies to
domain shift problems in regression and segmentation tasks and
to domain shift problems in non-parametric calibration methods such as conformal prediction.

\bibliographystyle{splncs04}
\bibliography{refs}
\end{document}